# MLRSNet: A Multi-label High Spatial Resolution Remote Sensing Dataset for Semantic Scene Understanding

Xiaoman Qi[a], Panpan Zhu[b,c], Yuebin Wang[a, *], Liqiang Zhang[c], Junhuan Peng[a], Mengfan Wu[c], Jialong Chen[a], Xudong Zhao[a], Ning Zang[a] and P. Takis Mathiopoulos[d]

X. Qi and P. Zhu contributed equally to this work, Corresponding author: Yuebin Wang (e-mail: xxgcdxwyb@163.com).

[a] School of Land Science and Technology, China University of Geosciences, Beijing 100083, China

[b] College of Computer Science and Technology, Chongqing University of Posts and Telecommunications, Chongqing 400065, China

[c] Beijing Key Laboratory of Environmental Remote Sensing and Digital Cities, Faculty of Geographical Science, Beijing Normal University, Beijing 100875, China

[d] Department of Informatics and Telecommunications, National and Kapodestrian University of Athens, Athens 15784, Greece

**Abstract:** To better understand scene images in the field of remote sensing, multi-label annotation of scene images is necessary. Moreover, to enhance the performance of deep learning models for dealing with semantic scene understanding tasks, it is vital to train them on large-scale annotated data. However, most existing datasets are annotated by a single label, which cannot describe the complex remote sensing images well because scene images might have multiple land cover classes. Few multi-label high spatial resolution remote sensing datasets have been developed to train deep learning models for multi-label based tasks, such as scene classification and image retrieval. To address this issue, in this paper, we construct a multi-label high spatial resolution remote sensing dataset named MLRSNet for semantic scene understanding with deep learning from the overhead perspective. It is composed of high-resolution optical satellite or aerial images. MLRSNet contains a total of 109,161 samples within 46 scene categories, and each image has at least one of 60 predefined labels. We have designed visual recognition tasks, including multi-label based image classification and image retrieval, in which a wide variety of deep learning approaches are evaluated with MLRSNet. The experimental results demonstrate that MLRSNet is a significant benchmark for future research, and it complements the current widely used datasets such as ImageNet, which fills gaps in multi-label image research. Furthermore, we will continue to expand the MLRSNet. MLRSNet and all related materials have been made publicly available at

https://data.mendeley.com/datasets/7j9bv9vwsx/1 and https://github.com/cugbrs/MLRSNet.git.

## 1. Introduction

With the availability of enormous numbers of remote sensing images produced by satellites and airborne sensors, high-resolution remote sensing image analyses have stimulated a flood of interest in the domain of remote sensing and computer vision (Toth and Jóźków, 2016), such as image classification or land cover mapping (Cheng et al., 2017; Gómez et al., 2016; You and Dong, 2020; Zhao et al., 2016), image retrieval (Wang et al., 2016), and object detection (Cheng et al., 2014; Han et al., 2014), etc. The great potential offered by these platforms in terms of observation capability poses great challenges for semantic scene understanding (Bazi, 2019). For instance, as these data are obtained from different locations, at different times and even with different satellites or airborne sensors, there are large variations among the scene images, which creates difficulties for the tasks of semantic scene understanding, such as multi-label based image retrieval and image classification.

Furthermore, remote sensing images usually contain abundant information about ground objects, which creates challenges for semantic scene understanding tasks (Chaudhuri et al., 2017). But it is extremely expensive for labeling each piece of data accurately when the amount of data is huge. Therefore, some research on weakly-supervised segmentation based on image-level using the information of multi-label classification network has attracted the attention of some scholars (Ge et al., 2018; Xia et al., 2015). Moreover, there have been many explorations in the use of multi-label data, such as land cover classification (Stivaktakis et al., 2019), high-precision image retrieval (Chaudhuri et al., 2017), image semantic segmentation (Xia et al., 2015), or migrate the model of multi-label data training to other visual tasks (*e.g.*, image object recognition) (Gong et al., 2019). Therefore, multi-label datasets now attract increasing attention in the remote sensing community owing to that they are not expensive but have a lot of research potential. For these reasons, multi-label annotation of an image is necessary to present more details of the image and improve the performance of scene understanding. In addition, the multi-label annotation of an image can produce potential correlations among the labels, such as "road" and "car" tend to occur synchronously in a remote sensing image, and "grass" and "water" often accompany "golf course". This will provide a better understanding of scene images, which is impossible for single-label image scene understanding. Therefore, annotating images with multiple labels is a vital step for semantic scene understanding in remote sensing.

What is more, previous studies have proven that traditional machine learning methods cannot adequately mine ground object scene information (Cordts et al., 2016; Jeong et al., 2019; Kendall et al., 2017; Zhu et al., 2019). Recently, deep learning approaches, as a popular technology, have shown the great potential of providing solutions to problems related to semantic scene understanding, and many scholars have conducted relevant studies (Fang et al., 2020; Han et al., 2018; Hu et al., 2015; Ma et al., 2019; Paoletti et al., 2018; Wang et al., 2018; Zhang et al., 2016; Zhou et al., 2019). Such as, a highly reliable end-to-end real-time object detection-based situation recognition system was proposed for autonomous vehicles (Jeong et al., 2019). In another work (Cordts et al., 2016), the authors determined that fully convolutional networks achieve decent results in urban scene understanding. And scene classification CNNs were proved that they significantly outperform previous approaches (Zhou et al., 2017). In the reference (Workman et al., 2017), a novel CNN architecture for estimating geospatial functions, such as population density, land cover, or land use, was proposed. Moreover, CNNs were also used to identify weeds (Hung et al., 2014) and vehicles (Chen et al., 2014), etc.

Additionally, there exists a logarithmic relationship between the performance of deep learning methods on vision tasks and the quantity of training data used for representation learning was proven recently (Sun et al., 2017). This work demonstrated that the power of CNNs on large-scale image recognition tasks can be substantially improved if the CNNs are trained on large multi-perspective samples. At present, there exist some widely used various-scale annotated datasets, including image classification datasets like ImageNet (Deng et al., 2009), Places (Zhou et al., 2017), PASCAL VOC (Everingham et al., 2015), YouTube-8M (Abu-El-Haija et al., 2016), semantic segmentation datasets like PASCAL Context (Mottaghi et al., 2014), Microsoft COCO (Lin et al., 2014), Cityscapes (Cordts et al., 2016) and Mapillary Vistas Dataset (Neuhold et al., 2017). However, in these benchmarks, the data of outdoor objects on the ground are usually collected from ground-level views. In addition, the object-centric remote sensing image datasets constructed for scene classification, for instance, AID (Xia et al., 2017), NWPU-RESISC45 (Cheng et al., 2017), the Brazilian coffee scene dataset (Penatti et al., 2015), the UC-Merced dataset (Yang and Newsam, 2010), and WHU-RS19 dataset (Xia et al., 2010). But these datasets are insufficient to understand the scene due to the high intra-class diversity and low inter-class variation, with the limited number of remote sensing images (Xia et al., 2017). The SEN12MS dataset (Schmitt et al., 2019) attracts more attention in the domain of land use mapping recently. It consists of 180,662 triplets sampled over all meteorological seasons. Each triplet concludes a dual-pol synthetic aperture radar (SAR) image patches, a multi-spectral Sentinel-2 image patches, and four different

MODIS land cover maps following different internationally established classification schemes. However, the SEN12MS contains no more than 17 classes under a selected classification schemes, which may be also insufficient for understanding the complex real world.

Moreover, it is worth noting that each image in most of the afore-mentioned datasets is annotated by a single label representing the most significant semantic content of the image. However, single-label annotation is sufficient for simple problems, such as distinguishing between coffee class and noncoffee class but is difficult to address more complex scene understanding tasks. Multiple label-related methods have recently been found to be useful for scene understanding, such as multi-label image search and retrieval problems, where multiple class labels are simultaneously assigned to each image (Boutell et al., 2004; Li et al., 2010; Ranjan et al., 2015; Zhang and Zhou, 2007). Thus, several published multi-label archives are publicly available, for example, multi-label UAV image datasets such as the Trento dataset (Bazi, 2019) and the Civezzano dataset (Bazi, 2019) and multi-label RSIR archives such as MLRSIR (Shao et al., 2018). The Trento dataset and the Civezzano dataset both contain 14 classes and contain a total of 4,000 images and 4,105 images, respectively. A multi-label remote sensing image retrieval (RSIR) archive was released in 2017, which is considered to be the first open-source dataset for multi-label RSIR (Chaudhuri et al., 2017). Afterward, MLRSIR (Shao et al., 2018), which is a pixel-wise dataset for multi-label RSIR, was presented by Wuhan University and has a total number of 21 broad categories with 100 images per category. However, training the CNNs using the above datasets easily results in overfitting since the CNN models used for multi-label archives often contain millions of parameters. Thus, a considerable quantity of labeled data will be required to fully train the models. Although BigEarthNet (Sumbul et al., 2019) can deal with the problem of overfitting, the limitation in the distribution and the unique data source could reduce the interclass diversity, which raises difficulties for developing robust scene understanding algorithms.

To overcome the above issues and better understand ground objects, in this paper, we propose a novel large-scale high-resolution multi-label remote sensing dataset termed "MLRSNet" for semantic scene understanding. It contains 109,161 high-resolution remote sensing images that are annotated into 46 categories, and the number of sample images in a category varies from 1,500 to 3,000. The images have a fixed size of $256 \times 256$ pixels with various pixel resolutions. Moreover, each image in the dataset is tagged with several of 60 predefined class labels, and the number of labels associated with each image varies from 1 to 13. Moreover, we illustrate the construction procedure of the MLRSNet dataset and give evaluations and comparisons of several deep learning methods for multi-label based image classification and image retrieval. The experiments indicate that multi-label based deep learning methods can achieve better performance on image classification and image retrieval.

In summary, three major contributions of this paper are as follows:

(1) A review of related popular datasets is provided by giving a summary of their properties. Covering different scale single-label datasets and multi-label datasets, of which most are usually insufficient for remote sensing scene understanding tasks.

(2) A multi-label high spatial resolution remote sensing dataset, *i.e.*, MLRSNet is developed for semantic scene understanding. To our knowledge, the dataset is a large high-resolution multi-label remote sensing dataset with the most abundant multi-label information. And the dataset has high intraclass diversity, which can provide a better data resource for evaluating and advancing the numerous methods in semantic scene understanding areas.

(3) The state-of-the-art neural network methods for multi-label image classification and multi-label image retrieval using MLRSNet are evaluated. These results show that deep-learning-based methods achieve significant performance for multi-label based image classification and image retrieval tasks.

## 2. MLRSNet: A Multi-label High Spatial Resolution Remote Sensing Dataset

How to improve the performance of existing multi-label image classification and retrieval approaches using machine learning and other artificially intelligent technologies has attracted much attention in the remote sensing community (Chua et al., 2009). However, for learning-based methods, a large number of labeled samples are required. To advance the state-of-art methods in scene understanding of remote sensing, we construct the MLRSNet, a new large-scale high-resolution multi-label remote sensing image dataset.

### 2.1 Description of MLRSNet

MLRSNet is composed of 109,161 labeled RGB images from all around the world annotated into 46 broad categories: *airplane, airport, bareland, baseball diamond, basketball court, beach, bridge, chaparral, cloud, commercial area, dense residential area, desert, eroded farmland, farmland, forest, freeway, golf course, ground track field. harbor&port, industrial area, intersection, island, lake, meadow, mobile home park, mountain, overpass, park, parking lot, parkway, railway, railway station, river, roundabout, shipping yard, snowberg, sparse residential area, stadium, storage tank, swimming pool, tennis court, terrace, transmission tower, vegetable greenhouse, wetland, and wind turbine*. The number of sample images varies greatly with different broad categories, from 1,500 to 3,000, as shown in Fig. 1. Additionally, each image in the dataset is assigned several of 60 predefined class labels, and the number of labels associated with each image varies between 1 and 13. The number of images present in the dataset associated with each predefined label is listed in Table 1, and some samples with corresponding multi-label results are shown in Fig. 2.

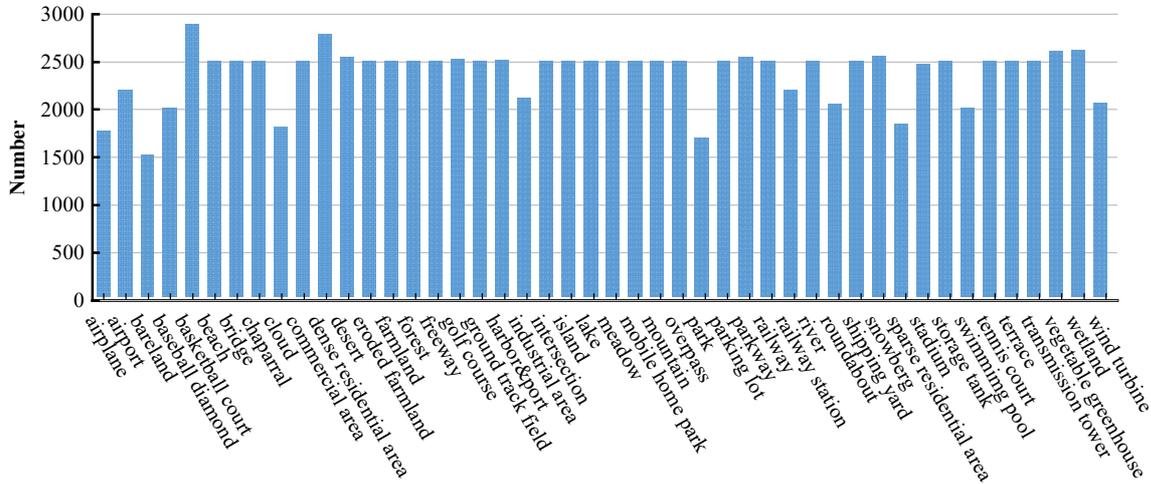

Fig. 1. Illustration of the number of samples per category in MLRSNet. There are 109,161 samples within 46 scene categories.

Table 1. Number of images present in the dataset for each class label. There are 60 predefined class labels in total.

| Class label | Number | Class label | Number | Class label | Number |
|---|---|---|---|---|---|
| airplane | 2,306 | freeway | 2,500 | roundabout | 2,039 |
| airport | 2,480 | golf course | 2,515 | runway | 2,259 |
| bare soil | 39,345 | grass | 49,391 | sand | 11,014 |
| baseball diamond | 1,996 | greenhouse | 2,601 | sea | 4,980 |
| basketball court | 3,726 | gully | 2,413 | ships | 4,092 |
| beach | 2,485 | habor | 2,492 | snow | 3,565 |
| bridge | 2,772 | intersection | 2,497 | snowberg | 2,555 |
| buildings | 51,305 | island | 2,493 | sparse residential area | 1,829 |
| cars | 34,013 | lake | 2,499 | stadium | 2,462 |
| chaparral | 5,903 | mobile home | 2,499 | swimming pool | 5,078 |
| cloud | 1,798 | mountain | 5,468 | tanks | 2,500 |
| containers | 2,500 | overpass | 2,652 | tennis court | 2,499 |
| crosswalk | 2,673 | park | 1,682 | terrace | 2,345 |
| dense residential area | 2,774 | parking lot | 7,061 | track | 3,693 |
| desert | 2,537 | parkway | 2,537 | trail | 12,376 |
| dock | 2,492 | pavement | 56,383 | transmission tower | 2,500 |
| factory | 2,667 | railway | 4,399 | trees | 70,728 |
| field | 15,142 | railway station | 2,187 | water | 27,834 |
| football field | 1,057 | river | 2,493 | wetland | 3,417 |
| forest | 3,562 | road | 37,783 | wind turbine | 2,049 |

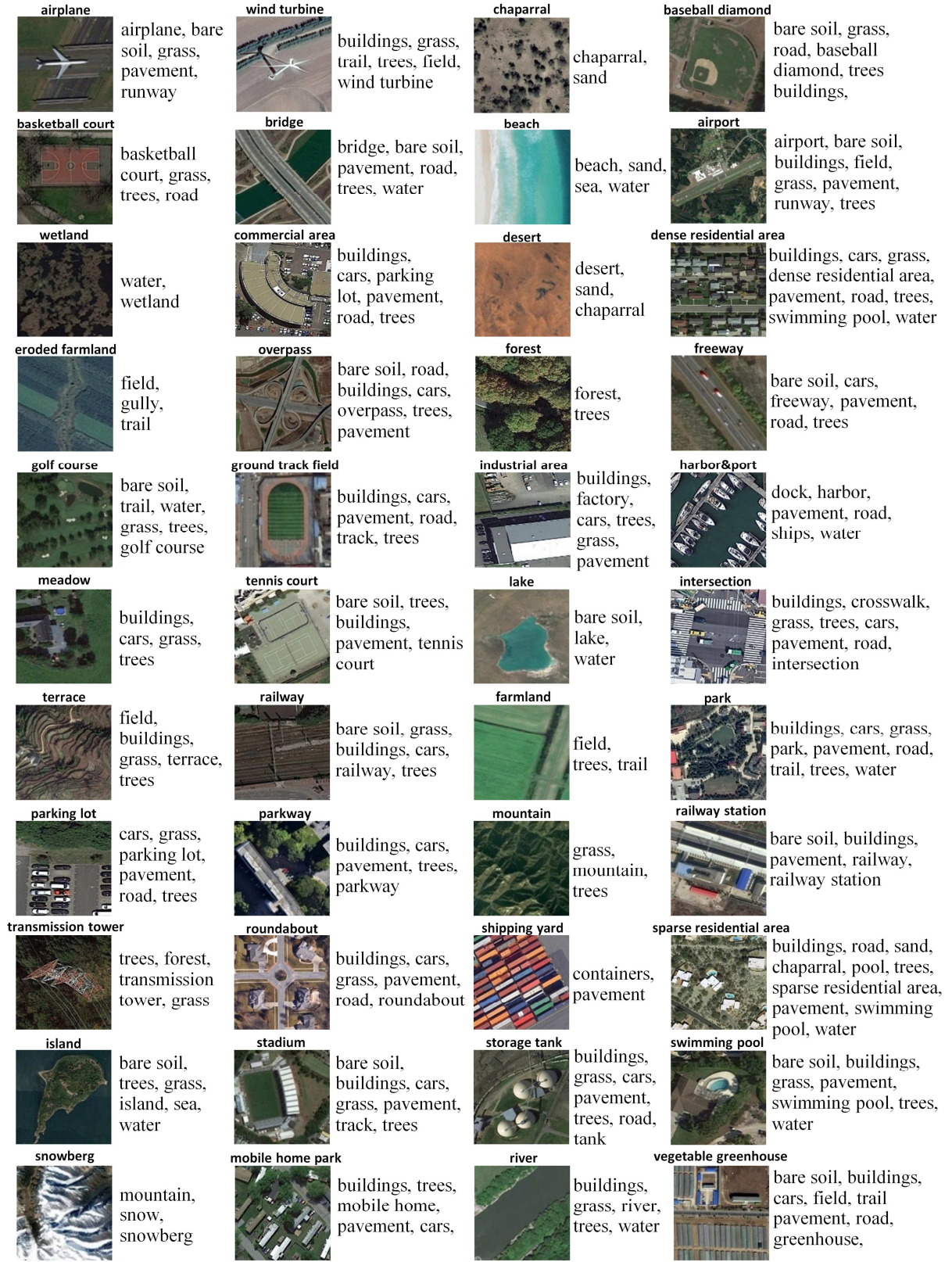

Fig. 2. Example images of 44 categories (except "bareland" and "cloud") from the MLRSNet dataset are shown, and the corresponding multi-labels of each image are reported at the right of the related image.

Table 2. Statistics of our database and comparisons of current state-of-the-art remote sensing benchmarks.

| Dataset | Number of Total Samples | Number of Categories | Sample Number in Each Category | Image sizes | Image Spatial Resolution (m) | Reference |
|---|---|---|---|---|---|---|
| UC-Merced | 2,100 | 21 | 100 | 256×256 | 0.3 | (Everingham et al., 2015) |
| NWPU-RESISC45 | 31,500 | 45 | 700 | 256×256 | ~30 to 0.2 | (Cheng et al., 2017) |
| AID | 10,000 | 30 | 220-420 | 600×600 | 8 to 0.5 | (Xia et al., 2017) |
| MLRSIR | 2,100 | 21 | 100 | 256×256 | 0.3 | (Shao et al., 2018) |
| SEN12MS | 564,768 | 17 | - | 256×256 | 10 to 500 | (Schmitt et al., 2019) |
| BigEarthNet | 590,326 | 43 | 328-217,119 | up to 120×120 | 10, 20, 60 | (Sumbul et al., 2019) |
| **MLRSNet** | **109,161** | **46** | **1,500-3,000** | **256×256** | **~10 to 0.1** | **our work** |

*The number of categories for SEN12MS is counted following the International Geosphere Biosphere Programme (IGBP) classification scheme (Loveland and Belward, 1997).*

Besides, MLRSNet has multi-resolutions: the pixel resolution changes from about 10 meters to 0.1 meters, and the size of each multi-label image is fixed to 256×256 pixels to cover a scene with various resolutions. Compared to the afore-mentioned scene understanding datasets in Section 1, MLRSNet has a more significantly large variability in terms of geographic origins and number of object categories. Different from ImageNet (Deng et al., 2009), which collects the data of outdoor objects from ground-level views, MLRSNet describes the objects on Earth from an overhead perspective through satellite or aerial sensors. Therefore, deep neural networks can be trained based on MLRSNet combined with ImageNet. We can achieve much higher recognition precision of the scene and effectively address the challenges of object rotation, within-class variability, and between-class similarity. Table 2 lists the differences between MLRSNet and other widely used scene understanding datasets.

In contrast, with the existing remote sensing image datasets, MLRSNet has the following notable characteristics:

**Hierarchy:** MLRSNet contains 3 first-class categories, such as land use and land cover (*e.g.*, commercial area, farmland, forest, industrial area, mountain), natural objects and landforms (*e.g.*, beach, cloud, island, lake, river, chaparral), as well as man-made objects and landforms (*e.g.*, airplane, airport, bridge, freeway, overpass), 46 second-class categories (as shown in Fig. 1) and 60 third-class labels (as shown in Table 1).

**Multi-label:** As shown in Fig. 2, each image in the MLRSNet dataset has one or more corresponding labels because the remote sensing image usually contains many classes of objects that are not mutually exclusive. Several experiments (Shao et al., 2018; Zhang et al., 2018) have indicated that multi-label datasets tend to achieve more satisfactory performance than single-label datasets in the tasks of image classification or image retrieval.

**Large-scale:** As shown in Table 2, MLRSNet has a large number of high-resolution multi-label remote sensing scene images. It contains 109,161 high-resolution remote sensing images annotated into 46 categories, and the number of sample images in a category varies from 1,500 to 3,000, all of which are larger than most other listed datasets. MLRSNet is a large-scale high-resolution remote sensing dataset collected for scene image recognition that can cover a much wider range of satellite or aerial images. It is meant to serve as an alternative to advance the development of methods in scene image recognition, particularly deep-learning-based approaches that require large quantities of labeled training data.

**Diversity:** To increase the generalization ability of the dataset, we attempt to characterize MLRSNet according to the object distributions for geographical and seasonal distribution, weather conditions, viewing perspectives, capturing time, and image resolution, *i.e.*, large variations in spatial resolution, viewpoint, object pose, illumination and background as well as occlusion.

## 2.2 Construction of MLRSNet

MLRSNet is a remote sensing community-led dataset for people who want to visualize the world with overhead perspectives. To construct the MLRSNet, we gather a team of more than 50 annotators in the remote sensing domain and spend more than six months for the whole process. The construction of MLRSNet is mainly composed of three procedures, *i.e.*, scene sample collection, database quality control, and database sample diversity improvement.

### 2.2.1 Scene Category and Sample Collection

To satisfy **hierarchy** criteria, the first asset of a high-quality dataset is covering an exhaustive list of representative scene categories. To achieve this goal, we investigate all scene classes of the existing datasets to form a list of scene categories. In the process, we merge some similar semantic scene categories in different datasets into a new category. For example, "playground" and "ground track field" are taken as "ground track field", and "harbor" and "port" are taken as a new category called "harbor&port". We also search the keywords "object-based image analysis (OBIA)", "geographic object-based image analysis (GEOBIA)", "land cover classification", "land use classification", "geospatial image retrieval" and "geospatial object detection" on Web of Science and Google Scholar to carefully select some new meaningful scene classes. Consequently, we obtain 46 scene categories in total, as shown in Fig.

1.

Moreover, most of existing dataset are labeled with the name of categories, which describe the most significant semantic content of the image, but the primitive classes (multiple labels) presented in images are ignored. Thus, we associate each image with one or more land-cover class labels (*i.e.*, multi-labels) based on visual inspection. For every scene category, we randomly select 100 images and annotate the primitive classes in the image. Next, we count the primitive classes in the image and filter out the primitive classes whose number is no more than 5. Finally, we get 60 multiple labels occurred frequently in remote sensing samples. Generally, scene categories are scene-level labels and primitive classes are object-level labels.

Compared with other satellite or aerial image datasets, the samples in MLRSNet have more additional meaningful information, such as ***hierarchy*** and ***multi-label*** information. Particularly, when having multiple labels of scene samples, by comparing the sample features, we can search the ground object more precisely. With this information, many multi-label tasks can be solved, such as multi-label image classification, multi-label image retrieval and object detection.

Data ***diversity*** is ensured by data sources and manual control. We collect data samples by more than 20 people from multi-resolution, multi-continent, multi-time, multi-light and multi-viewpoint data sources to characterize MLRSNet. Like most of the existing datasets, such as AID (Xia et al., 2017), NWPU-RESISC45 (Cheng et al., 2017), and WHU-RS19 (Xia et al., 2010), MLRSNet is also extracted from Google Earth where images are from different remote imaging sensors. The satellite sensors include but are not limited to the GeoEye-1, WorldView-1, WorldView-2, SPOT-7, Pleiades-1A, and Pleiades-1B. And images also can be collected by the cameras for aerial photography. We collect data from all over the world to satisfy the diversity criteria, and the samples in MLRSNet cover more than 100 countries and regions. In addition, we control the data diversity. More details can be found in Section 2.2.3.

## 2.2.2 Data Quality Control

Aiming to develop a highly accurate dataset, we implement a quality control process. In the process, we rely on another 20 annotators to verify the ground-truth label of each candidate image collected in the previous process, including scene sample annotation, counting confidence score, and disposing of confusing data.

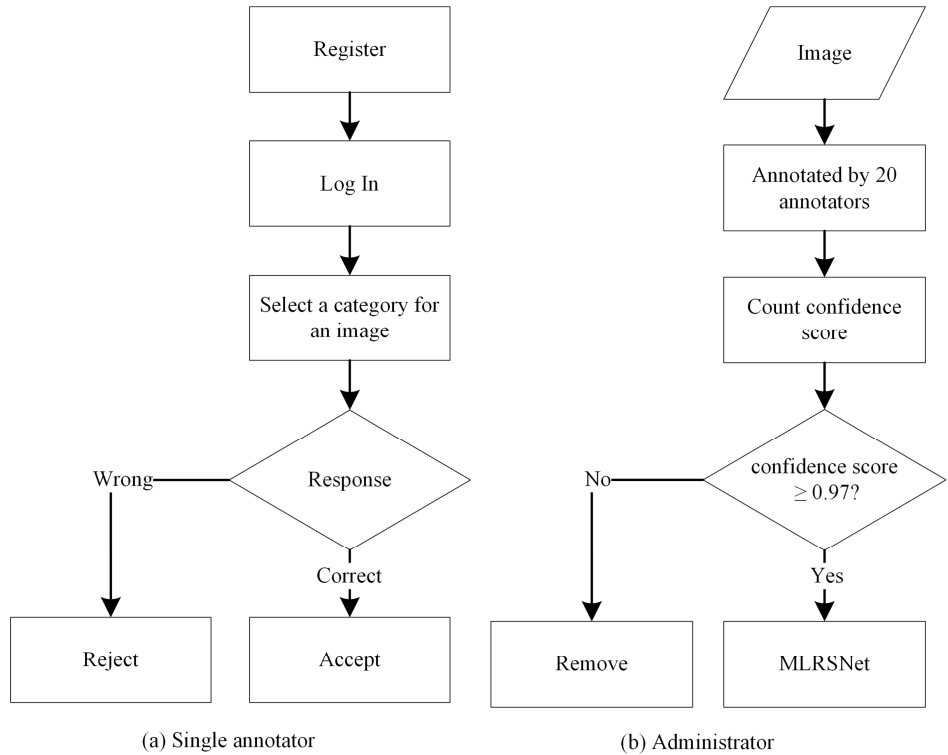

(a) Single annotator                 (b) Administrator

Fig. 3. The experimental process of quality control. (a) The experimental process of a single annotator. (b) The experimental process of the administrator.

Fig. 3 illustrates the experimental process of quality control. A sample is randomly presented to the annotator for selecting a category from 46 predefined category names. If the annotation result remains consistent with the ground-truth label of each candidate sample, the system gives an "accept" response; otherwise, it gives a "reject" response. Because of the subjectivity of human and the complexity of the image, different images need a different number of annotations. The solution, according to ImageNet (Deng et al., 2009), is to require multiple annotators to tag the images individually. While annotators are instructed to label an image, we make a confidence table (see Table 3) to dynamically determine the number of annotations needed for different categories of images. Table 3 shows examples for "airport", "bridge", "island" and "parkway". The confidence score indicates the probability of an image is a good image given the annotator votes. After data labeling for approximately two months, every sample is labeled several times until a predetermined confidence score threshold is reached. Therefore, the data samples with a confidence score $\geq$ 0.97 are retained while others are removed.

Table 3. Confidence score table for different categories of data samples, showing how annotators' judgment influence the probability of an image being a good image.

| Accept | Reject | airport | bridge | island | parkway |
|--------|--------|---------|--------|--------|---------|
| 0 | 1 | 0.13 | 0.05 | 0.03 | 0.14 |
| 1 | 0 | 0.80 | 0.87 | 0.89 | 0.67 |
| 1 | 1 | 0.51 | 0.49 | 0.50 | 0.52 |
| 2 | 0 | 0.90 | 0.97 | 0.98 | 0.84 |
| 0 | 2 | 0.05 | 0.03 | 0.02 | 0.13 |
| 3 | 0 | 0.97 | 0.99 | 1.00 | 0.90 |
| 2 | 1 | 0.82 | 0.86 | 0.88 | 0.73 |

We observe that the boundaries of some data pairs are blurry, *e.g.*, airport and runway, intersection and crosswalk, desert and bareland (see Fig. 4). For this reason, we gather our annotators for a discussion about the boundary of these data pairs. After that, we begin a second labeling round for data in these ambiguous pairs with a confidence score of < 0.97. Similarly, after the second labeling round, we preserve samples scored ⩾ 0.97 and deprecate others. Finally, we collect more than 100,000 data samples within 46 scene categories.

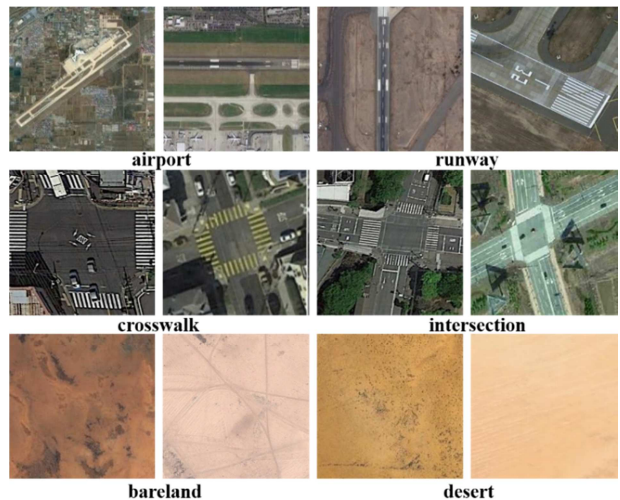

Fig. 4. Boundaries among scene categories can be blurry. The images show a soft transition between airport vs. runway, crosswalk vs. intersection and bareland vs. desert.

### 2.2.3 Data Diversity Improvement

An ideal dataset, expected to generalize well, should have high diversity, which means it should include a high variability of appearances, locations, resolutions, scales and background clutter and occlusions.

We use a measure to quantify the relative diversity of image datasets referred to as the practice in reference (Zhou et al., 2017). During the procedure of comparing the diversity of data samples, while our dataset diversity is lower than other datasets, we can improve the quality and number of data samples for a certain category.

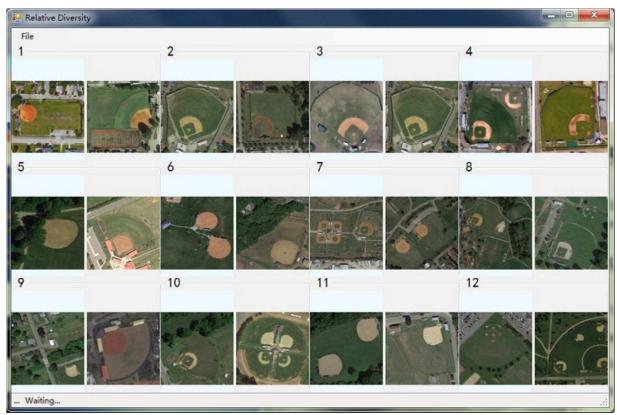

Fig. 5. A screenshot of the visual tool that computes the relative diversity of scene datasets. Different pairs of samples were randomly presented to a person who was instructed to select the most similar pair. Each trial was composed of 4 pairs from each database, giving a total of 12 pairs to choose from.

We develop a tool with a graphical user interface, as shown in Fig. 5. We ask 10 annotators to measure the relative diversities among AID (Xia et al., 2017), NWPU-RESISC45 (Cheng et al., 2017), and MLRSNet. Each time, different pairs of samples are randomly presented to an annotator who is instructed to select the most similar pair. Each trial is composed of 4 pairs from each database, giving a total of 12 pairs to choose from. We run 50 trials per category and 10 observers per trial, for the 20 categories in common.

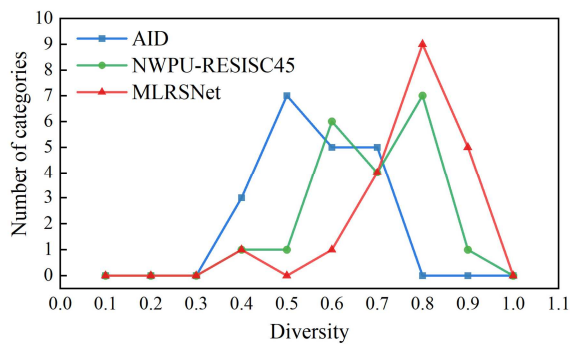

Fig. 6. Relative diversity of each category (20 categories) in a different dataset. MLRSNet (in red line) contains the most diverse set of images.

Fig. 6 shows the results of the relative diversity for all 20 scene categories common to the three databases. The results show that there is a large variation in terms of diversity among the three datasets, and MLRSNet is the most diverse of the three datasets. The average relative diversity on each dataset is for MLRSNet 0.78, for AID (Xia et al., 2017) is 0.56, and for NWPU-RESISC45 (Cheng et al., 2017) is 0.69. The categories with the smallest variation in diversity in MLRSNet are baseball diamond, beach, sparse residential area and storage tank. Then, we conduct a random rotation, resize and crop for all images in these four categories.

## 3. Scene Classification

Scene classification is a fundamental task in remote sensing image understanding. Recently, classification using convolutional neural networks (CNNs) has achieved significant performance. MLRSNet can be taken as a benchmark to evaluate the classification performances of different CNNs.

### 3.1 Experimental Settings

Eight popular CNN architectures, *i.e.*, InceptionV3 (Szegedy et al., 2016), VGGNet16 (Simonyan and Zisserman, 2014), VGGNet19 (Simonyan and Zisserman, 2014), ResNet50 (He et al., 2016), ResNet101 (He et al., 2016), DenseNet121 (Huang et al., 2017), DenseNet169 (Huang et al., 2017) and DenseNet201 (Huang et al., 2017) are chosen to address the remote sensing image classification problem, and the details of each model are shown in Table 4. It should be noticed that the final layer of CNNs is replaced by a dense connection with 60 nodes, and the result of dense connection is activated by a sigmoid function. Sigmoid maps the values of output vector of the network to an interval of (0, 1) indicating the score for each class. Then we binarize the output vector with a threshold of 0.5 to generate a multi-label prediction, similar to [1, 0, 1, …, 0], where 1 indicates the image is annotated with the corresponding label and otherwise it is 0. Finally, the models are trained on MLRSNet. We call the fine-tuned models MLRSNet-CNN, *i.e.*, MLRSNet-VGGNet16.

Table 4. Details of each CNNs model be used in the experiment

| CNNs | Layers | Parameters | Top-1 Accuracy | Top-5 Accuracy | year |
|------|--------|------------|----------------|----------------|------|
| InceptionV3 | 47 | 23M | 0.779 | 0.937 | 2014 |
| VGGNet16 | 16 | 138M | 0.713 | 0.901 | 2014 |
| VGGNet19 | 19 | 143M | 0.713 | 0.900 | 2014 |
| ResNet50 | 50 | 25M | 0.749 | 0.921 | 2015 |
| ResNet101 | 101 | 44M | 0.764 | 0.928 | 2015 |
| DenseNet121 | 121 | 8M | 0.750 | 0.923 | 2017 |
| DenseNet169 | 169 | 14M | 0.762 | 0.932 | 2017 |
| DenseNet201 | 201 | 20M | 0.773 | 0.936 | 2017 |

*The top-1 and top-5 accuracy refers to the model's performance on the ImageNet validation dataset.*

***InceptionV3*** (Szegedy et al., 2016)***:*** The inception module was first proposed in reference (Szegedy et al., 2016) by Google and was adopted for image classification and object detection in the ImageNet Large-Scale Visual Recognition Challenge 2014 (ILSVRC14). The main hallmark of this architecture is the improved utilization of the computing resources inside the network. However, the author explored methods to scale up networks in ways that aim at utilizing the added computation as efficiently as possible by suitably factorized convolutions and aggressive regularization, which is named InceptionV3.

***VGGNet16, VGGNet19*** (Simonyan and Zisserman, 2014)**:** VGG was originally developed for the ImageNet dataset by the Oxford Visual Geometry Group in the ILSVRC14. To investigate the effect of the convolutional network depth on its accuracy in the large-scale image recognition setting, Simonyan and Zisserman thoroughly evaluated the network of increasing depth using an architecture with very small ($3\times3$) convolution filters, which showed a significant improvement in the accuracies. In this work, we use two models that show the corresponding performance in scene classification, named VGGNet16 and VGGNet19.

***ResNet50，ResNet101*** (He et al., 2016)**:** Residual Nets (ResNet) is a framework presented by Microsoft Research to ease the training of networks that are substantially deeper than those used previously. This model won the 1st place on the ILSVRC 2015 classification task. ResNet50 is the 50-layer ResNet, and ResNet101 is the 101-layer ResNet.

***DenseNet121, DenseNet169, DenseNet201*** (Huang et al., 2017)**:** The dense convolution network (DenseNet) connects each layer to other layers in a feed-forward manner and has $\dfrac{L(L+1)}{2}$ direct connections for convolutional networks with *L* layers. DenseNets are widely used because they have several compelling advantages, such as alleviating the vanishing gradient problem, strengthening feature propagation, encouraging feature reuse and substantially reducing the number of parameters. DenseNet121, DenseNet169 and DensesNet201 are the 121-layer, 169-layer and 201-layer DenseNet, respectively.

To comprehensively evaluate the classification performances of different CNNs, three training-testing ratios are considered: (i) 20%-10%-70%, *i.e.*, we randomly select 20% of the dataset for training, 10% for validation and others for testing; (ii) 30%-10%-60%; (iii) 40%-10%-50%.

We choose TensorFlow and the Python package Keras for our experiments. The aforementioned eight methods, which are pretrained on the ImageNet dataset, are obtained from the URL: https://github.com/fchollet/deep-learning-models/releases . In the experiment, binary cross-entropy measures how far away from the true value (which is either 0 or 1) the prediction is for each of the classes and then averages these class-wise errors to obtain the final loss. The formula of binary cross-entropy adopted for multi-label classification can be shown as follow:

$$L = \frac{1}{mq}\sum_{i=1}^{m}\sum_{j=1}^{q}(t_i^j \log \hat{t}_i^j + (1-t_i^j)\log(1-\hat{t}_i^j)) \qquad (1)$$

where $t_i^j \in (0,1)$ denotes the *j*th ground-truth label for training image $X_i$. $\hat{t}_i^j$ is the output of the sigmoid layer. *m* is the number of training images and *q* is the number of classes in total.

To improve the generalization capability, we fine-tune the models by using the parameters listed in Table 5. All the CNN models are implemented on a 2.10 GHz 48-core CPU. In addition, a TITAN RTX GPU is used for acceleration. We try to avoid introducing random errors by duplicating experiments. In this study, we repeat the experiment five times and plot an error-bar graph by counting the results.

Table 5. Parameters utilized for model fine-tuning.

| Package | Epochs | Batch Size | Optimizer | Learning rate |
|---------|--------|------------|-----------|---------------|
| Keras   | 10     | 32         | Adam      | 0.01          |

## 3.2 Evaluation Protocols

We compute two commonly used evaluation metrics, *i.e.*, mean average precision and average $F_1$ score to quantitatively evaluate the classification results.

The $F_1$ score is a comprehensive metric for evaluating classification performance for each model and can be defined as:

$$F_1 = \frac{precision \times recall}{precision + recall} \times 2 \tag{2}$$

where $precision = \frac{|L_c \cap L_r|}{|L_c|}$ and $recall = \frac{|L_c \cap L_r|}{|L_r|}$. $L_c$ is the final label vector of the networks for a sample. $L_r$ is the ground-truth label of the sample. $\cap$ denotes intersection. $|\cdot|$ denotes the number of nonzero entries. A higher value of the $F_1$ score represents a better classification performance. Here, it should be pointed out that all precision and recall values are computed separately for each sample and then averaged across samples.

## 3.3 Experimental Results

We evaluate eight methods using MLRSNet and present the results of multi-label image classification as follows. As shown in Table 6 and Table 7, the fine-tuned models can achieve good classification performances on MLRSNet, which indicates that deep-learning-based models have the ability to obtain discriminative features. The statistical results of repeated trials are shown in Fig. 7 and Fig. 8.

It is worth noting that MLRSNet-DenseNet201 obtains significantly better metric values in the comparative experiment. In particular, the MLRSNet-DenseNet201 model can achieve an overall improvement as the number of data points increased. MLRSNet-DenseNet201 and MLRSNet-DenseNet169 achieve over 0.80 $F_1$ score in the 10 epochs when the training ratio is 20%.

Table 6. Mean Average Precision (%) of the eight fine-tuned models under different training ratios.

| CNNs | Training ratios | | |
|---|---|---|---|
| | 20% | 30% | 40% |
| MLRSNet-InceptionV3 | 81.50 | 82.33 | 84.84 |
| MLRSNet-VGGNet16 | 67.88 | 72.66 | 75.39 |
| MLRSNet-VGGNet19 | 66.12 | 69.53 | 73.60 |
| MLRSNet-ResNet50 | 82.65 | 84.28 | 86.01 |
| MLRSNet-ResNet101 | 83.26 | 84.19 | 85.72 |
| MLRSNet-DenseNet121 | 75.96 | 77.99 | 80.25 |
| MLRSNet-DenseNet169 | 82.16 | 86.42 | 87.35 |
| MLRSNet-DenseNet201 | **87.25** | **87.84** | **88.77** |

*The best results are shown in bold.*

Table 7. $F_1$ score of the eight fine-tuned models under different training ratios.

| CNNs | Training ratios | | |
|---|---|---|---|
| | 20% | 30% | 40% |
| MLRSNet-InceptionV3 | 0.7746 | 0.8016 | 0.8146 |
| MLRSNet-VGGNet16 | 0.5743 | 0.6534 | 0.6855 |
| MLRSNet-VGGNet19 | 0.5677 | 0.6120 | 0.6329 |
| MLRSNet-ResNet50 | 0.7530 | 0.8176 | 0.8353 |
| MLRSNet-ResNet101 | 0.7618 | 0.7703 | 0.8226 |
| MLRSNet-DenseNet121 | 0.7154 | 0.7389 | 0.7571 |
| MLRSNet-DenseNet169 | 0.8138 | 0.8408 | 0.8521 |
| MLRSNet-DenseNet201 | **0.8381** | **0.8414** | **0.8538** |

*The best results are shown in bold.*

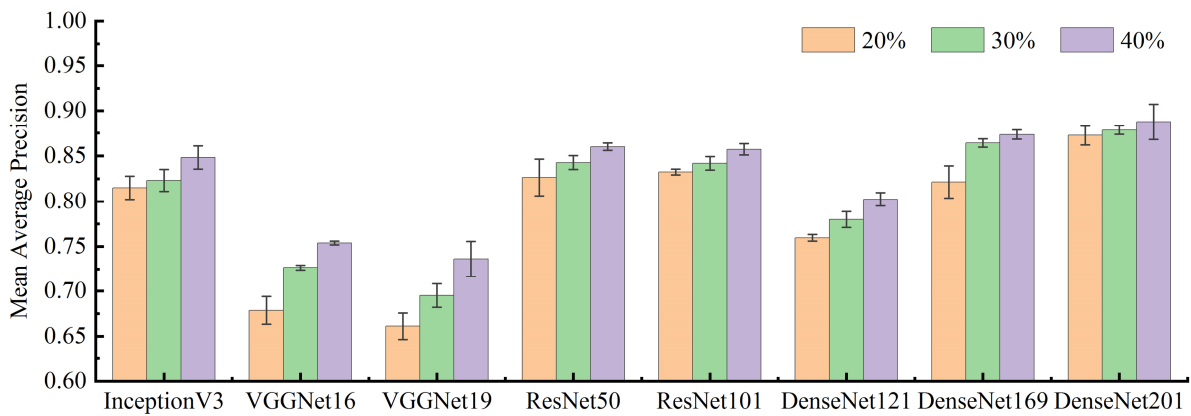

Fig. 7. The statistical results of Mean Average Precision. The bar chart shows the average, and the error line presents the standard deviation.

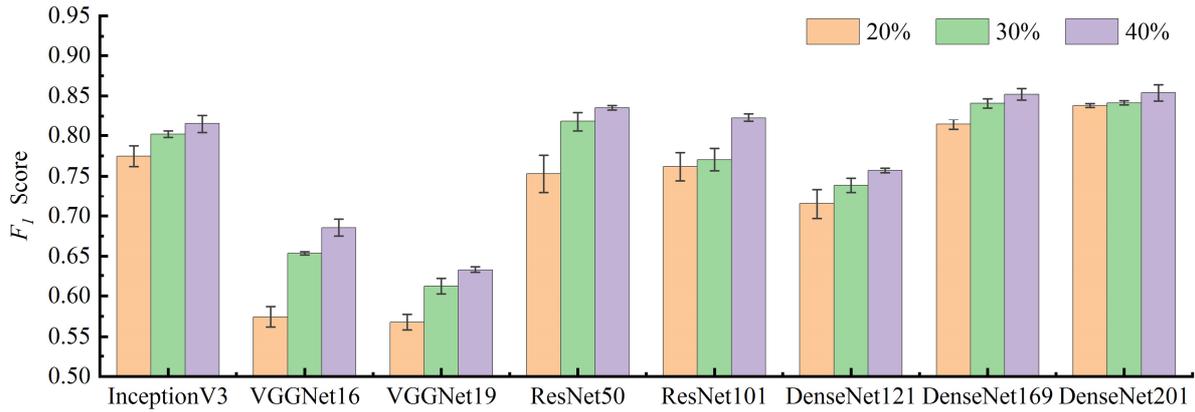

Fig. 8. The statistical results of $F_1$ score. The bar chart shows the average, and the error line presents the standard deviation.

Moreover, with increasing training data, the performances of the models increase. This suggests that increasing the data size can further improve the performances of deep learning models.

Fig. 9 shows several annotation examples on MLRSNet when fine-tuned DenseNet201 is employed. Here, annotations with black font are included in ground-truth labels, whereas annotations with red font are incorrect labels tagged by the model. The green fonts are correct labels, but the model does not tag them. It is obvious that the first seven images are all annotated correctly. The 8[th] to 10[th] images do not include incorrect annotations but neglecting no more than two correct labels, and the last five images contain one incorrect label. Hence, we can further find that MLRSNet-DenseNet201 has outstanding performance in multi-label classification. This model also can be adopted to help us to label the scene images in the future.

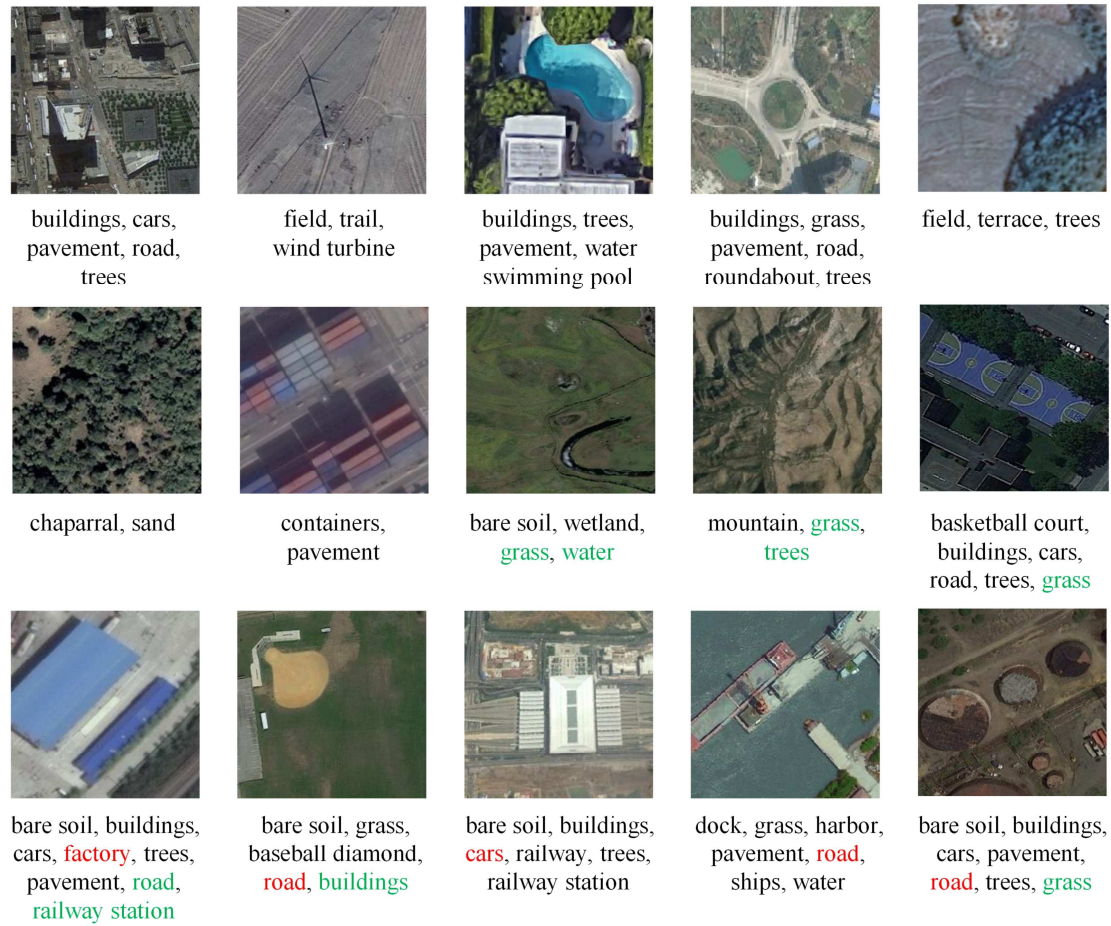

Fig. 9. Some samples of image classification by the MLRSNet-DenseNet201 model. Multi-labels of each image are reported below the related image, and the red font indicates the incorrect classification result, while the green font indicates the correct labels, but the model is not tagged.

## 4. Image Retrieval

With the sharp increase in the volume of remote sensing images, image retrieval has become an important topic of research in RS. We then show the application of MLRSNet to image retrieval. Similarly, it can be used as a benchmark dataset to evaluate the retrieval performances of different models.

### 4.1 Experimental Settings

We use the afore-mentioned eight fine-tuned CNN models (in Section 3) to evaluate the retrieval performances of the dataset in this work. Images are fed into CNNs, and the features are extracted from the last layer of each network. Euclidean distance is selected to calculate the similarity between the query images and images in the retrieval archives. Only if the retrieval results and the query image belong to the same category can we considered that the query has been satisfied.

Since the fine-tuned CNNs which are trained with the 40% images of MLRSNet are applied to perform the retrieval experiment, the remaining 60% images of MLRSNet are adopted as the testing queries and retrieval database. In order to make full use of the dataset for retrieval experiments, and minimize the random error, in this section, we randomly split MLRSNet into testing queries and retrieval database at three different ratios: 5% vs. 55%, 10% vs. 50%, and 15% vs. 45%. Taking 5% vs. 55% for example, 5% of images from each category are selected as query images to query the rest of the dataset. Databases of different sizes are further used to validate the effectiveness of MLRSNet for image retrieval experiments. Table 8 shows the number of images in testing queries and the number of images in retrieval database when 5%, 10% and 15% of images from each category are selected as the testing queries.

Table 8. The number of images in testing queries and in retrieval database when different percentages are chosen.

| Percentage | 5% | 10% | 15% |
|---|---|---|---|
| Testing queries | 3,275 | 6,550 | 9,825 |
| Retrieval database | 62,222 | 58,947 | 55,672 |

## 4.2 Retrieval Metrics

To evaluate the retrieval performance, we use average normalized modified retrieval rank (ANMRR), mean average precision (*mAP*), and precision at k (*P@k*, where *k* is the number of retrieved images) as metrics. In the following experiments, the ANMRR, mAP, and *P@k* are the averaged values over all the query images.

For a detailed description of ANMRR, we refer the reader to the reference (Manjunath et al., 2001). The formulas of the last two metrics are as follows.

$$P@k = \frac{1}{N_q} \sum_{i=1}^{N_q} \frac{N_s}{K} \tag{3}$$

$$mAP = \frac{1}{N_q} \sum_{i=1}^{N_q} \frac{1}{m_i} \sum_{k=1}^{m_i} p@k \tag{4}$$

where $N_q$ is the number of queries, $N_s$ represents for a given query the number of the images in the result that are considered as the correct image. $K$ is the number of retrieved images. $m_i$ is the number of result images for a given query $i$. We define $m_i = 100$, that is, we calculate the mean of top-100 average retrieved precision.

## 4.3 Experimental Results

The precision-recall curves of eight methods with different percentages of testing queries are shown in Fig. 10. Table 9 shows the values of ANMRR, *mAP*, and *P@k* (*k*=10, 50, 100, 500) obtained when MLRSNet-CNNs are used, and different percentages of the testing queries are chosen.

Table 9. The retrieval results of different methods. For ANMRR, lower value indicates better performance, while for *mAP* and *P@k*, larger is better.

| Percentage | Models | ANMRR | *mAP* | *P@10* | *P@50* | *P@100* | *P@500* |
|---|---|---|---|---|---|---|---|
| 5% | MLRSNet-Inception V3 | 0.3506 | 0.7307 | 0.7524 | 0.7270 | 0.7130 | 0.6575 |
| | MLRSNet-VGGNet16 | 0.5503 | 0.4868 | 0.5073 | 0.4833 | 0.4699 | 0.4107 |
| | MLRSNet-VGGNet19 | 0.6248 | 0.3690 | 0.3878 | 0.3656 | 0.3539 | 0.3084 |
| | MLRSNet-ResNet50 | 0.2407 | 0.8333 | 0.8462 | 0.8314 | 0.8220 | 0.7610 |
| | MLRSNet-ResNet101 | 0.3001 | 0.7580 | 0.7730 | 0.7556 | 0.7456 | 0.6928 |
| | MLRSNet-DenseNet121 | 0.3440 | 0.7513 | 0.7758 | 0.7478 | 0.7316 | 0.6577 |
| | MLRSNet-DenseNet169 | 0.1665 | 0.8874 | 0.8969 | 0.8863 | 0.8791 | 0.8414 |
| | MLRSNet-DenseNet201 | **0.1557** | **0.8959** | **0.9031** | **0.8949** | **0.8898** | **0.8596** |
| 10% | MLRSNet-Inception V3 | 0.3518 | 0.7321 | 0.7547 | 0.7283 | 0.7135 | 0.6535 |
| | MLRSNet-VGGNet16 | 0.5579 | 0.4721 | 0.4922 | 0.4683 | 0.4552 | 0.3956 |
| | MLRSNet-VGGNet19 | 0.6238 | 0.3709 | 0.3883 | 0.3676 | 0.3561 | 0.3070 |
| | MLRSNet-ResNet50 | 0.2496 | 0.8208 | 0.8339 | 0.8191 | 0.8082 | 0.7415 |
| | MLRSNet-ResNet101 | 0.3073 | 0.7435 | 0.7695 | 0.7394 | 0.7219 | 0.6832 |
| | MLRSNet-DenseNet121 | 0.3493 | 0.7604 | 0.7810 | 0.7571 | 0.7443 | 0.6423 |
| | MLRSNet-DenseNet169 | 0.1693 | 0.8834 | 0.8928 | 0.8823 | 0.8747 | 0.8327 |
| | MLRSNet-DenseNet201 | **0.1583** | **0.8936** | **0.9018** | **0.8925** | **0.8869** | **0.8520** |
| 15% | MLRSNet-Inception V3 | 0.3569 | 0.7225 | 0.7450 | 0.7189 | 0.7042 | 0.6372 |
| | MLRSNet-VGGNet16 | 0.5626 | 0.4533 | 0.4760 | 0.4489 | 0.4358 | 0.3847 |
| | MLRSNet-VGGNet19 | 0.6213 | 0.3694 | 0.3883 | 0.3661 | 0.3521 | 0.3053 |
| | MLRSNet-ResNet50 | 0.2503 | 0.8141 | 0.8299 | 0.8123 | 0.7988 | 0.7317 |
| | MLRSNet-ResNet101 | 0.3096 | 0.7516 | 0.7675 | 0.7498 | 0.7368 | 0.6745 |
| | MLRSNet-DenseNet121 | 0.3505 | 0.7355 | 0.7631 | 0.7306 | 0.7141 | 0.6301 |
| | MLRSNet-DenseNet169 | 0.1724 | 0.8786 | 0.8893 | 0.8775 | 0.8687 | 0.8223 |
| | MLRSNet-DenseNet201 | **0.1577** | **0.8935** | **0.9003** | **0.8929** | **0.8869** | **0.8465** |

*The best results are shown in bold.*

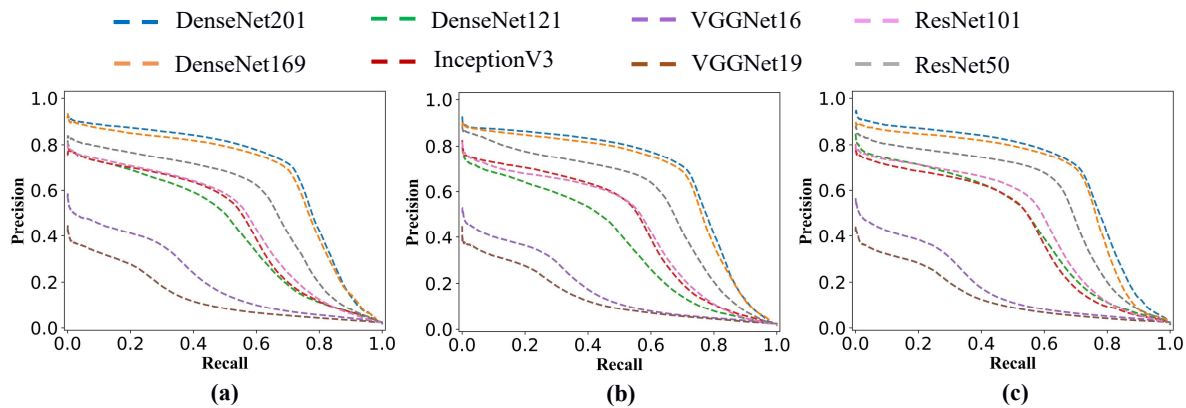

Fig. 10. The precision-recall curves of MLRSNet-CNNs with different percentages of the testing queries. (a), (b), (c) represent the percentages of testing queries are 5%, 10%, 15%, respectively.

By analyzing Fig. 10 and Table 9, it can be observed that most of the MLRSNet-CNNs obtain impressive retrieval results regardless of different percentages of the testing queries. It demonstrates that CNNs indeed perform well in the image retrieval task using MLRSNet. Especially, MLSNet-DenseNet201 is higher than other models with different percentages of testing queries in *mAP* and *P@k* (*k* = 10, 50, 100, 500) for the MLRSNet dataset. The *mAP* results of MLRSNet-DenseNet201 indicate a relative improvement of about 1% against MLRSNet-DenseNet169. But the *mAP* results of MLRSNet-DenseNet201 indicate a relative increase of 6.26% ~ 7.94% against MLRSNet-ResNet50. Meanwhile, MLSNet-DenseNet201 is lower than other models with different percentages of testing queries in ANMRR. In Fig. 10, the precision-recall curve of MLSNet-DenseNet201 is superior to other methods no matter the size of the retrieval database. As a result, the MLRSNet-DenseNet201 can achieve better performance compared with other approaches. The reason lies in that each layer in the DenseNet accepts all previous layers' features as input, which can maximize the information flow between all layers in the network. Besides, for the three types of DenseNet models, as the number of layer increases, the models can obtain more representative and discriminative image features, showing better experimental performances.

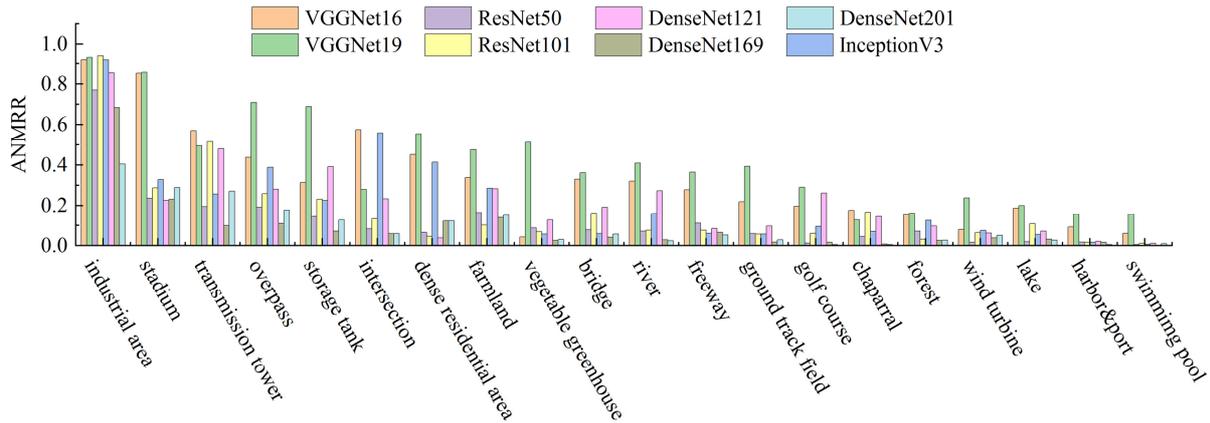

Fig. 11. The ANMRR results of features for a set of categories with 10% of images extracted by MLRSNet-CNNs.

Moreover, Fig. 11 shows the ANMRR results of features extracted from several categories by the eight methods when we choose 10% of the MLRSNet as the testing queries. It is evident that the ANMRR result of MLRSNet-VGGNets is large for most categories, meaning that its retrieval performance is poor. This may lie in that the simple network structures only can obtain limited representations of images, leading to poor experimental performances. Meanwhile, it can be seen that the same network performs differently for various categories. In particular, most networks can perform well in the categories that can be recognized easily (e.g., lake, harbor&port and swimming pool).

Taking MLRSNet-DenseNet201 as an example, we show the precision-recall curves of a set of categories when percentage of testing queries is 10% in Fig. 12. This illustrates that although the whole retrieval performance of MLRSNet-DenseNet201 is impressive, the performances on several categories are not satisfactory, *e.g.*, basketball court and commercial area. The reason may be that the high intra-class diversities of these two categories increase the difficulties of image retrieval.

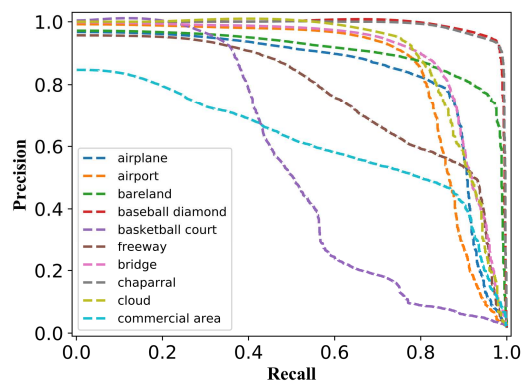

Fig. 12. The precision-recall curves of MLRSNet-DenseNet201 on a set of categories with the MLRSNet.

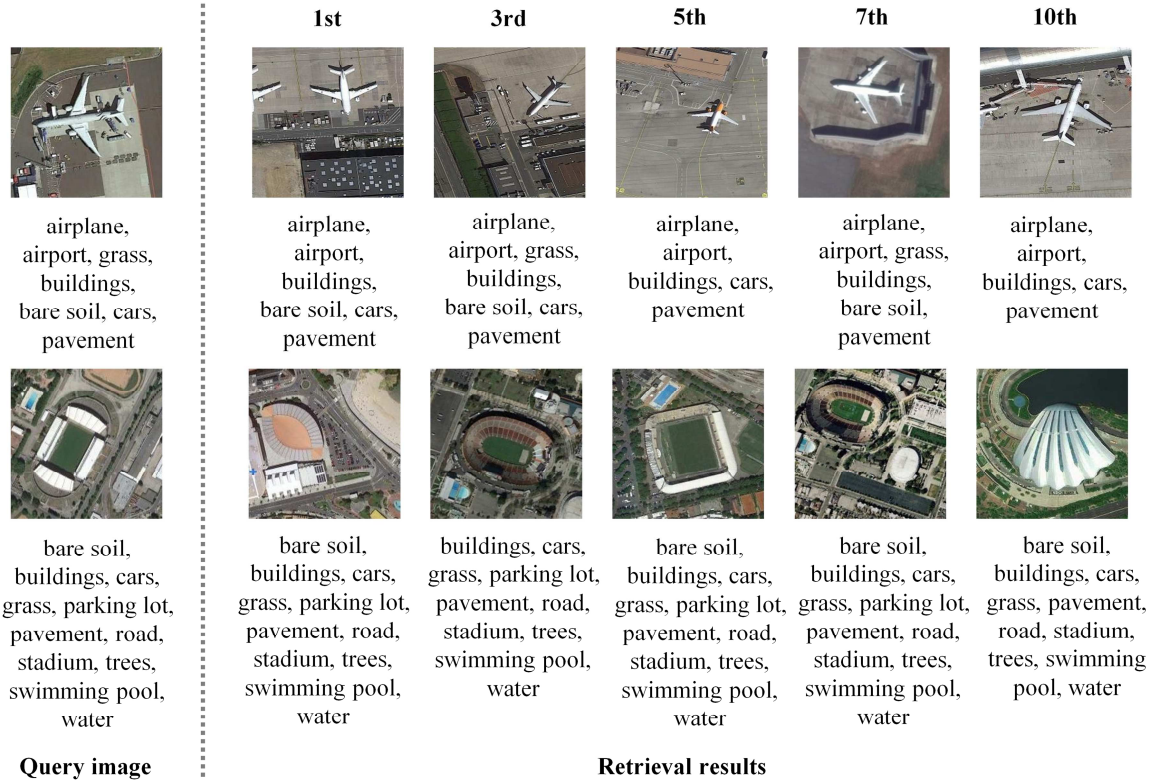

Fig. 13. The retrieval results of airplane (top) and stadium (bottom) categories by the MLRSNet-DenseNet201 model when the percentage of the testing queries is 10%.

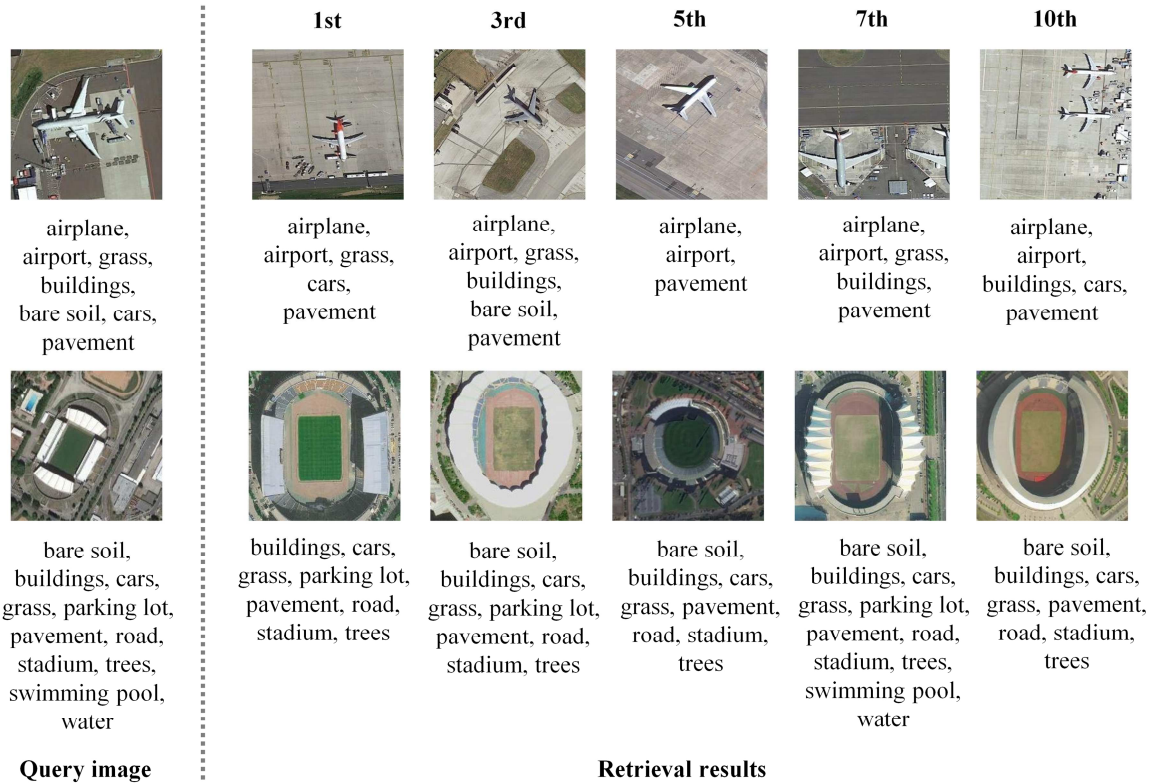

Fig. 14. The retrieval results of airplane (top) and stadium (bottom) categories by the SLRSNet-DenseNet201 model when the percentage of the testing queries is 10%.

Fig. 13 shows two examples of result images retrieved by the MLRSNet-DenseNet201 when the query image is randomly selected from the airplane category and the stadium category with the percentages of the testing queries is 10%. The multi-labels associated with the image are given below the related image. From the retrieved results, it is obvious that the MLRSNet-DenseNet201 model accurately detects the multi-label image objects associated with a given query image and retrieve the most visually similar images from the database.

Compared with single-label remote sensing image retrieval, image retrieval using a multi-label dataset can add more restricted conditions (multi-label) to the retrieval process, thereby achieving more accurate image retrieval results. We use the name of the category as the label for each image in the category, and then apply the single-label remote sensing dataset (SLRSNet) to train the DenseNet201 model and perform the same retrieval experiments. The retrieval results are shown in Fig. 14. We can find that the multi-label image retrieval results contain more common labels with the retrieval image by comparing Fig. 13 and Fig. 14. It indicates that the multi-label image retrieval results are more similar to the retrieval image. Taking the stadium category as an example, when we query stadium images that contain a "swimming pool", the multi-label image retrieval better matches the labels "swimming pool" and "water", yet the single-label image retrieval has difficulty meeting this requirement.

## 5. Conclusion

The MLRSNet is a multi-label high spatial resolution remote sensing dataset for semantic scene understanding with deep learning from the overhead perspective. MLRSNet has distinctive characteristics: hierarchy, large-scale, diversity, and multi-label. Experiments, including multi-label scene classification and image retrieval, are conducted with different deep neural networks. From these experimental results, we can conclude that MLRSNet can be adopted as a benchmark dataset for performance evaluations of multi-label image retrieval and scene classification. MLRSNet complements the current large object-centric datasets such as ImageNet. In future work, we will continue to expand the MLRSNet and apply the dataset to other recognition tasks, such as semantic/instance segmentation in large-scale scene images and ground object recognition.